\documentclass[times,twocolumn,final,authoryear]{elsarticle}

\usepackage{ycviu}
\usepackage{framed,multirow}

\usepackage{amssymb}
\usepackage{latexsym}
\usepackage{amsmath}

\usepackage{url}
\usepackage{xcolor}
\definecolor{newcolor}{rgb}{.8,.349,.1}

\journal{Computer Vision and Image Understanding}

\begin{document}

\thispagestyle{empty}

\clearpage

\ifpreprint
  \setcounter{page}{1}
\else
  \setcounter{page}{1}
\fi

\begin{frontmatter}

\title{Physics-based Shading Reconstruction for Intrinsic Image Decomposition}

\author[1]{Anil S. \snm{Baslamisli}\corref{cor1}} 
\cortext[cor1]{Corresponding author: 
}
\ead{a.s.baslamisli@uva.nl}
\author[2]{Yang \snm{Liu}}
\author[2]{Sezer \snm{Karaoglu}}
\author[1,2]{Theo \snm{Gevers}}

\address[1]{University of Amsterdam, Science Park 904, 1098XH Amsterdam, the Netherlands}
\address[2]{3DUniversum, Science Park 400, 1098XH Amsterdam, the Netherlands}

\received{1 May 2013}
\finalform{10 May 2013}
\accepted{13 May 2013}
\availableonline{15 May 2013}
\communicated{S. Sarkar}

\begin{abstract}
We investigate the use of photometric invariance and deep learning to compute intrinsic images (albedo and shading). We propose albedo and shading gradient descriptors which are derived from physics-based models. Using the descriptors, albedo transitions are masked out and an initial sparse shading map is calculated directly from the corresponding $RGB$ image gradients in a learning-free unsupervised manner. Then, an optimization method is proposed to reconstruct the full dense shading map. Finally, we integrate the generated shading map into a novel deep learning framework to refine it and also to predict corresponding albedo image to achieve intrinsic image decomposition. By doing so, we are the first to directly address the texture and intensity ambiguity problems of the shading estimations. Large scale experiments show that our approach steered by physics-based invariant descriptors achieve superior results on MIT Intrinsics, NIR-RGB Intrinsics, Multi-Illuminant Intrinsic Images, Spectral Intrinsic Images, As Realistic As Possible, and competitive results on Intrinsic Images in the Wild datasets while achieving state-of-the-art shading estimations.
\end{abstract}

\begin{keyword}
\MSC 41A05\sep 41A10\sep 65D05\sep 65D17
\KWD Keyword1\sep Keyword2\sep Keyword3

%% MSC codes here, in the form: \MSC code \sep code
%% or \MSC[2008] code \sep code (2000 is the default)
\end{keyword}

\end{frontmatter}

%\linenumbers

%% main text
\section{Introduction}
\label{sec:intro}
Intrinsic image decomposition is the inverse problem of recovering the image formation components, such as reflectance and shading~\citep{Barrow1978}. The shading component consists of light effects such as direct illumination, geometry, shadow casts and ambient light. The reflectance component represents the (albedo) color of an object and is free of any lighting effect. Intrinsic images are favorable for various computer vision tasks. For example, albedo images are beneficial for semantic segmentation algorithms because of their illumination invariant representation~\citep{Baslamisli2018ECCV}. Similarly, most of the scene editing applications, such as recoloring, rely on albedo images~\citep{Ye2014}, whereas shading images are preferred for relighting tasks~\citep{Shu2017}.

The pioneering work on intrinsic image computation is the Retinex algorithm by \citet{Land1971} which uses a heuristic that is based on the rectilinear Mondrian world assumption. In a Mondrian world, where surfaces have piece-wise constant colors, strong gradients correspond to albedo changes, while shading variations are related to weaker ones. Then, using a re-integration algorithm (i.e. Poisson) over the strong (albedo) gradients, the albedo component is computed. However, classifying image gradients into albedo or shading is not a trivial task due to various photometric effects such as strong shadow casts, illuminant color, surface geometry changes or weak albedo transitions. For instance, shadow boundaries or abrupt changes in surface geometry may cause strong intensity shifts and may therefore be interpreted as albedo changes. Moreover, the Mondrian world assumption do not apply to real world scenes. Other traditional approaches usually utilize an optimization process by introducing constraints on the intrinsic components~\citep{Gehler2011,Shen2011,Barron2015}. Most of the priors aim at constraining the albedo component such as global reflectance sparsity, piece-wise constant reflectance or chromaticity reflectance correlation. On the other hand, the shading intrinsic is usually constrained by a smoothness prior. 

\begin{figure}[t]
\centering
\includegraphics[width=1\linewidth]{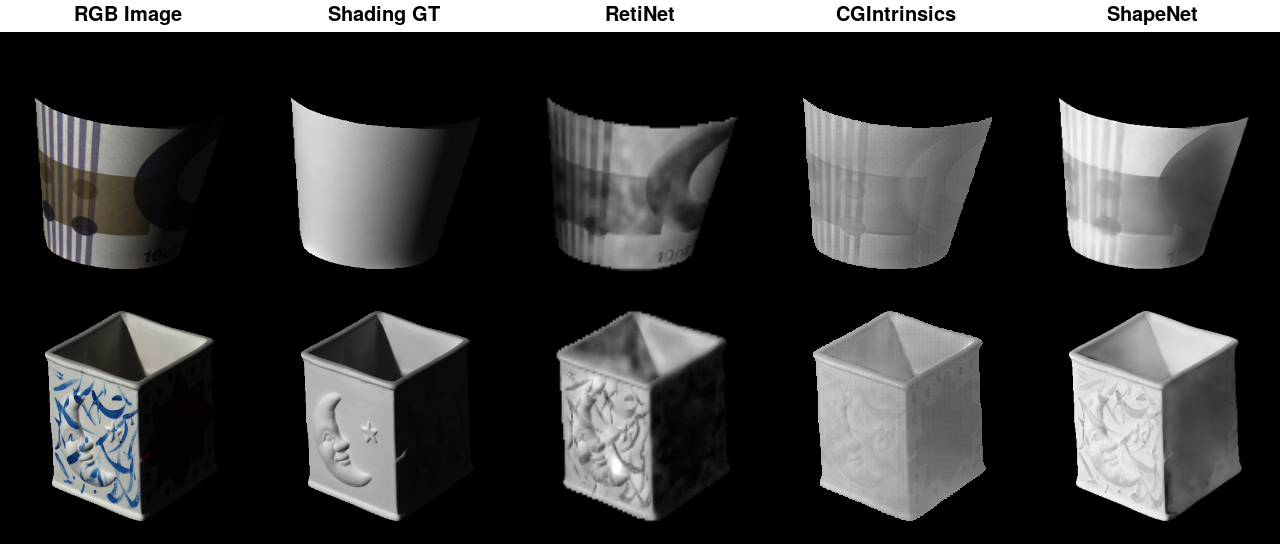}
   \caption{Color leakage problem in the estimated shading maps. It negatively effects the albedo separation from the shading.}
\label{fig:color_leakage}
\end{figure}

More recent methods rely on deep learning models, specialized loss functions, and large scale datasets. For example, \citet{Baslamisli2018CVPR} provide an end-to-end solution to the Retinex approach in a deep learning framework, \citet{Li2018ECCV} combine four datasets with specialized loss functions to impose constraints, and \citet{Lettry2018b} investigate adversarial learning. With the availability of densely annotated synthetic datasets and multiple constraints on the albedo component, CNN-based methods are capable of estimating high quality albedo maps. However, CNN-based shading estimations regularly suffer from texture and intensity ambiguities (e.g. albedo leakage) introducing (color) artifacts in the shading profiles. See Figure~\ref{fig:color_leakage} for an illustration.

In the early days of photometric invariance in computer vision, invariant image descriptors were widely used for different vision tasks. These descriptors are invariant to certain image capturing conditions so that the vision algorithms are not affected by them, such as illumination color, surface geometry or camera position. Successful results were demonstrated for object recognition \citep{Gevers1997}, image retrieval \citep{Gevers2000}, and shadow removal \citep{Finlayson2006}. As CNN-based shading estimations suffer from  (color) artifacts, physics-based invariant features may be useful to steer the intrinsic image decomposition process.

Therefore, we investigate the use of photometric invariance and deep learning to compute intrinsic images (albedo and shading). We propose albedo and shading gradient descriptors which are derived from physics-based models. Using the descriptors, albedo transitions are masked out and an initial shading map is calculated directly from the corresponding $RGB$ image gradients in a learning-free manner (unsupervised). Then, an optimization method is proposed to reconstruct the full shading map. Finally, we integrate the shading map into a deep learning model to achieve full intrinsic image decomposition.

\textbf{Contributions.} \textit{1.} We are the first to use photometric invariance and deep learning to address the intrinsic image decomposition task. \textit{2.} We propose albedo and shading gradient descriptors using physics-based models. \textit{3.} The shading map is calculated directly from the corresponding $RGB$ image gradients in a learning-free (unsupervised) manner. \textit{4.} We propose a novel deep learning model to leverage the physics-based shading map for the intrinsic image decomposition task.  By doing so \textit{5.} we are the first to directly address the color leakage problem in the estimated shading maps. Finally, \textit{5.} we extend the dataset of \citet{Baslamisli2018CVPR} from 15,000 to 50,000 images to train our models, which will be publicly available.

\section{Related Work}
Intrinsic image decomposition is an ill-posed and under-constrained problem. The pioneering work is the Retinex algorithm by \citet{Land1971} based on the assumption that albedo changes cause large gradients, whereas shading variations result in smaller ones. In general, traditional approaches use different optimization processes to constrain the intrinsic components together with the Retinex heuristic. For example, \citet{Gehler2011} impose constraints on the global albedo sparsity. SIRFS estimates shape, chromatic illumination, albedo, and shading from a single image by applying seven different constraints on the intrinsic components \citep{Barron2015}. Intrinsic Images in the Wild (IIW) model combines commonly used priors together with a dense conditional random field \citep{Bell2014}. \citet{Shen2011} use optimization to constraint neighboring pixels having similar intensity values to have similar albedo values. \citet{Shen2008} exploit non-local texture cues by constraining distinct points with the same intensity-normalized textures to have the same albedo values. Furthermore, user interactions are investigated as additional priors to specify albedo values \citep{Bousseau2009,Shen2013}. Finally, image sequences of the same scene under varying illumination are used to impose constant albedo \citep{Weiss2001,Matsushita2004}. Most of the priors mentioned above are related to the albedo intrinsic. It is partially due to color information being more descriptive for robust computer vision algorithms \citep{Sande2009}. It is also relatively harder to define priors for the shading intrinsic, because geometry and lighting information are entangled in the representation.

With the introduction of large-scale synthetic datasets, recent research use convolutional neural networks \citep{Shi2017,Baslamisli2018ECCV,Li2018ECCV}. \citet{Narihia2015} are the first to use CNNs to learn the task end-to-end in a data-driven manner. \citet{Shi2017} make use of a very large scale dataset along with a specialized network to exploit correlations between the intrinsic components. \citet{Baslamisli2018CVPR} convert the Retinex approach into a deep learning framework together with a physics-based image formation loss. \citet{Cheng2018} use a Laplacian pyramid inspired neural network architecture to exploit scale space properties. \citet{Lettry2018b} explore adversarial residual networks. \citet{Fan2018} apply a domain filter guided by a learned edge map to flatten the albedo estimations. \citet{Li2018ECCV} combine four datasets with specialized loss functions. \citet{Janner2017} explore the problem in a self-supervised setting by estimating albedo, shape, and lighting, where shape and lighting estimations are used to train a differentiable shading function. \citet{Baslamisli2020} further decomposes the shading into different photometric effects. Image sequences of the same scene under varying illumination are also explored by deep learning approaches \citep{Lettry2018,Li2018CVPR}. Recent work focusing on inverse rendering tasks also achieve superior albedo estimations \citep{Sengupta2019,Li2020}. Nonetheless, these methods are limited by indoor settings and require surface normal and environmental lighting supervision.

CNN-based methods are capable of estimating high quality albedo maps that are mostly free of photometric effects. However, their shading estimations are often negatively affected by albedo transitions causing texture ambiguities and intensity variations, as illustrated in Figure~\ref{fig:color_leakage}. To mitigate the problem, for example, \citet{Zhou2019} shift the problem of predicting shading to predicting surface normals and lighting properties. Yet, their work is limited by indoor settings and require additional modalities and supervision, similar to inverse rendering works. Another example is CGIntrinsics which over-smooths the shading estimations, yet that in return causes structure loss in the shading maps \citep{Li2018ECCV}.  As CNN-based shading estimations suffer from albedo artifacts, invariant image representations may be favorable to steer the process. They were widely used for various image understanding tasks \citep{Drew1998,Finlayson1995,Finlayson2006,Gevers1997,Gevers1998,Gevers2000}. One example is the illumination invariant color ratio features used for robust object recognition \citep{Finlayson1992}. \citet{Stricker1992} combine ratio histograms with boundary histograms for a more robust framework. \citet{Nayar1996} utilize color ratios for pose estimation. \citet{Matas1995} embed ratio information into a graph representation also for efficient object recognition. \citet{Barnard2000} identify probable shadow regions using color ratios. \citet{Gevers2001} exploit ratio gradients for image retrieval. As invariant image representations are independent of the certain imaging conditions, they may be useful to improve CNN-based shading estimation as part of intrinsic image decomposition. To this end, in this paper, we investigate the use of photometric invariance and deep learning to compute intrinsic images (albedo and shading).

\section{Methodology} 

\subsection{Image Formation Model}
\label{ss:ifm}
We use the dichromatic reflection model of \citet{Shafer1985} to describe an $RGB$ image. The model defines a surface (image) $I$ as a combination of diffuse $I_{d}$ and specular $I_{s}$ reflections as follows:
\begin{equation} \label{eq:shafer}
\begin{aligned}
&I = I_{d} + I_{s}\;.
\end{aligned}
\end{equation}
We assume that the diffuse reflection component dominates the imaging conditions and hence the effect of the specular reflection component is negligible, i.e. $I \approx I_{d}$. Then, an image \emph{I} over the visible spectrum $\omega$ is modelled by:
\begin{equation} \label{eq:imf}
I_{c} = m(\vec{n}, \vec{l}) \int_{\omega}^{} f_{c}(\lambda)\; e(\lambda)\; \rho(\lambda)\; \mathrm{d}\lambda \;,
\end{equation}
for three color channels $c \in \{R,G,B\}$, where $\vec{n}$ indicates the surface normal, $\vec{l}$ denotes the incoming light source direction, and \emph{m} is a function of the geometric dependencies (e.g. Lambertian $\vec{n} \cdot \vec{l}$). Furthermore, $\lambda$ represents the wavelength, $f$ indicates the camera spectral sensitivity, and $e$ describes the spectral power distribution of the light source. Finally, $\rho$ denotes the reflectance i.e. the albedo. Then, assuming a linear sensor response and narrow band filters ($f_c(\lambda_{c}$)), the equation can be simplified as follows:
\begin{equation} \label{eq:imf_mult}
I_{c} = m(\vec{n}, \vec{l}) \; e(\lambda_{c})\; \rho(\lambda_{c})\;=\;
m(\vec{n}, \vec{l}) \; e_c\; \rho_c\;.
\end{equation}
This equation models an image by the multiplication of its geometry $m(\vec{n}, \vec{l})^x$, albedo $\rho_c^x$ and light source properties $e_c^x$ at pixel $x$. Then, these characteristics are used to define intrinsic images as follows:
\begin{equation} \label{eq:iid}
I_{c}^x = S_{c}^x \times R_{c}^x \;,\;\;\;\;\\
S_{c}^x = m(\vec{n}, \vec{l})^x \; e_{c}^x\;,\;\;\;\;\\
R_{c}^x = \rho_{c}^x\;,
\end{equation}
\noindent where an image $I_{c}$ at $x$ can be modelled by the element-wise product of its shading $S_{c}$ and albedo $R_{c}$ components. If the light source $e$ is colored, then the color information is embedded in the shading component. 

\subsection{Albedo Gradients}
\label{ss:invariance}
\begin{figure}[t]
\includegraphics[width=1\linewidth]{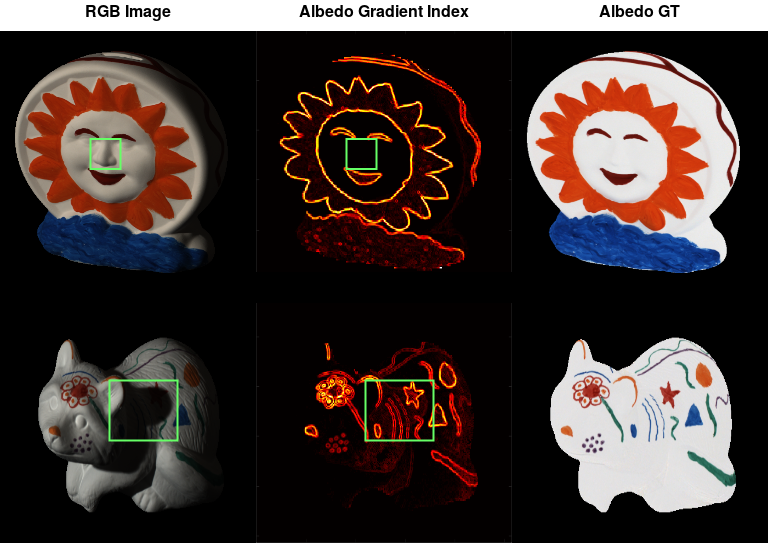}
   \caption{Finding albedo changes (edges) by the use of the albedo gradient index. Brighter values indicate a higher degree of albedo changes. Uniformly colored patches have low scores. Note the similarity of the albedo gradient index and the albedo ground-truth image. The \emph{sun} object shows invariance to geometry and strong shading, and the \emph{raccoon} object demonstrates invariance to shadows.}
\label{fig:uniform_color_index}
\end{figure}

Using Equation~\ref{eq:imf_mult}, the image formation model for the three color channels $c \in \{R,G,B\}$ becomes:
\begin{equation} \label{eq:cc}
\begin{aligned}
R^x = m(\vec{n}, \vec{l})^x \; e_{R}^x\; \rho_{R}^x\;, \\
G^x = m(\vec{n}, \vec{l})^x \; e_{G}^x\; \rho_{G}^x\;, \\
B^x = m(\vec{n}, \vec{l})^x \; e_{B}^x\; \rho_{B}^x\;.
\end{aligned}
\end{equation}
Considering only neighbouring pixels $x_1$ and $x_2$, locally constant illumination can be assumed: $e_{c}^{x_1} = e_{c}^{x_2}$ \citep{Land1971}.
By taking the difference of the logarithmic transformation of each color channel, the albedo descriptors are defined as follows:
\begin{equation} \label{eq:F}
m_1 = \Delta \log{\frac{R}{G}}\;,\;
m_2 = \Delta \log{\frac{R}{B}}\;,\;
m_3 = \Delta \log{\frac{G}{B}}\;.
\end{equation}
We illustrate the invariant properties of these albedo descriptors by plugging Equation~\eqref{eq:cc} into Equation~\eqref{eq:F} for $m_1$ as follows (same also holds for $m_2$ and $m_3$):
\begin{equation} \label{eq:demo}
\begin{aligned}
m_1 &= \Delta \log{\frac{R}{G}} = \log \frac{R^{x_1}}{G^{x_1}} - \log \frac{R^{x_2}}{G^{x_2}}\\  &=
(\log R^{x_1} - \log G^{x_1}) - ( \log R^{x_2} - \log G^{x_2})\\ &=
(
(\log m(\vec{n}, \vec{l}))^{x_1} + 
\log e_R^{x_1} + 
\log \rho_R^{x_1})
-
(\log m(\vec{n}, \vec{l}))^{x_1} + 
\log e_G^{x_1} \\ &+ 
\log \rho_G^{x_1})
)
\;-\;
(
(\log m(\vec{n}, \vec{l}))^{x_2} + 
\log e_R^{x_2} + 
\log \rho_R^{x_2})
-
(\log m(\vec{n}, \vec{l}))^{x_2}\\ &+ 
\log e_G^{x_2} + 
\log \rho_G^{x_2})
)\\ &=
\log \frac{ \rho_R^{x_1}} { \rho_G^{x_1} }
-
\log \frac{ \rho_R^{x_2}} { \rho_G^{x_2} }
=
\Delta \log \frac{  \rho_R} { \rho_G }\;,
\end{aligned}
\end{equation}
\noindent
where the remaining factor is only the albedo difference between two channels. The albedo change is a measure that is invariant to surface geometry $\vec{n}$, illumination direction $\vec{l}$, and its intensity and color $e$. If there is no albedo change (homogeneously colored patch), then the difference is zero. Sensor artifacts or noise may slightly deviate the value from zero. Therefore, the index can be used to identify regions with constant albedo. On the other hand, when the difference deviates significantly from zero, it corresponds to a true albedo change. Hence, this measure encodes spatial information of an image emphasizing on (illumination invariant) albedo edges. Then, we propose the \textit{albedo gradient index} as follows:
\begin{equation} \label{eq:agi}
\begin{aligned}
AGI = \sqrt{(\Delta \log \frac{R}{G})^2 + (\Delta \log \frac{R}{B})^2 + (\Delta \log \frac{G}{B})^2}\;.
\end{aligned}
\end{equation}
We calculate the albedo gradients over a local neighbourhood (patch) by using derivative filters (e.g. the derivative of a 2D Gaussian or Laplacian) to identify the changes. As a result, the average response of the albedo gradients is calculated. A neighborhood with a higher albedo gradient index value indicates a stronger albedo change, which is also illustrated in Figure~\ref{fig:uniform_color_index}. A patch with a constant index yields the homogeneous regions. The albedo gradient index is very intuitive and realized in real time. It is computed for a small threshold to remove possible problems caused by sensor artifacts and noise. 

\subsection{Shading Gradients}
So far, we have described that the albedo gradient index can be used to identify uniformly colored (homogeneous) patches. In a color image, if the pixel values share the same albedo, then the only source causing those pixel values to change is the shading component. For constant $\rho$ (satisfying $AGI \approx 0$) over an image neighborhood, the shading gradient can be computed by taking the difference of the logarithmic transformation of each color channel. We illustrate it on the red channel as follows (same also holds for green and blue channels):
\begin{equation} \label{eq:sci}
\begin{aligned}
\Delta \log R &= 
(\log m(\vec{n}, \vec{l})^{x_1} + \log e_R^{x_1} + \log \rho_R^{x_1})\\ &- (\log m(\vec{n}, \vec{l})^{x_2} + \log e_R^{x_2} + \log \rho_R^{x_2})\\ &=
\log m(\vec{n}, \vec{l})^{x_1} - \log m(\vec{n}, \vec{l})^{x_2} =
\Delta \log m(\vec{n}, \vec{l}).
\end{aligned}
\end{equation} 
\noindent
Note that it is only applied on the homogeneous patches. Logarithms are usually preferred to avoid numerical instabilities, yet note that also the derivatives of the $RGB$ channels can be taken to yield the following shading gradient index:
\begin{equation} \label{eq:shadingcomponent}
\begin{aligned}
SGI = \sqrt{(\Delta R)^2 + (\Delta G)^2 + (\Delta B)^2}\;.
\end{aligned}
\end{equation}
\noindent
Similar to the albedo gradient index, the average response is calculated, which results in representing the gradient field of an $RGB$ image. Note that (non-colored) shadows are included in the shading difference component i.e. when $e_R^{x_1} \neq e_R^{x_2}$. 

\subsection{Shading}
\begin{figure}[h]
\centering
\includegraphics[width=1\linewidth]{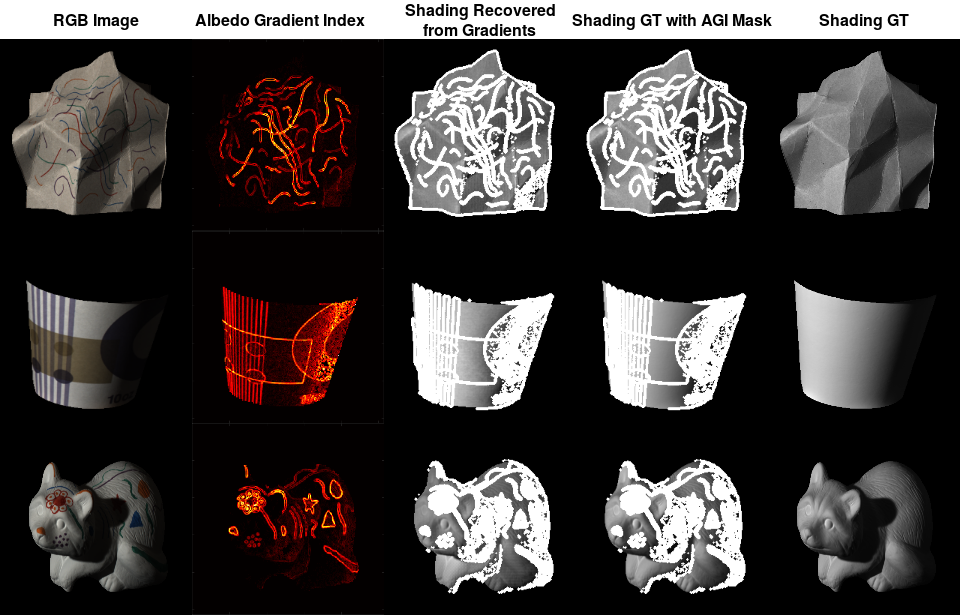}
   \caption{AGI-assisted physics-based shading gradient index. The albedo gradient index is directly computed from the $RGB$ image. Then, it is used to calculate a shading map by masking out regions that have albedo changes. The same mask is applied to the shading GT to show the resemblance.}
\label{fig:shading_prior}
\end{figure}

After obtaining the shading gradient, we reconstruct the shading map from its shading gradient fields. We use a publicly available algorithm to compute the global least squares reconstruction \citep{Harker2008,Harker2011}. Note that the albedo gradient index is used to detect uniformly colored (homogeneous) patches first. Then, the shading gradients are calculated only on the homogeneous patches. As a result, the reconstructed shading map is computed directly from the shading gradient fields of an $RGB$ image in an unsupervised manner. Since it is computed only on the homogeneous image regions (satisfying $AGI \approx 0$), a sparse shading map is obtained. Therefore, the representation is not affected by the albedo changes. The process is illustrated in Figure~\ref{fig:shading_prior}. In the end, we can generate a sparse shading map that is directly computed from the RGB image that is also very close to the ground-truth representation.

Then, a shading smoothness constraint is used to fill in the gaps based on the neighboring pixel information. To achieve that, we adapt a publicly available optimization framework that is originally designed for the depth completion task \citep{Zhang2018}. We modify the model to impose the shading smoothness constraint to achieve a full (dense) shading map. The objective function ($E$) is defined as the sum of squared errors with two terms $E = E_D + E_S$ as follows:  
\begin{equation} \label{eq:optimization}
\begin{aligned}
&E_D = \sum_{x \in T_{obs}}^{} || S(x) - S_0(x) ||^2\;, \\
&E_S = \sum_{p, q \in N}^{} || S(p) - S(q) ||^2\;, \\
\end{aligned}
\end{equation}
where $T_{obs}$ denotes the pixels that are available (not empty) in the initial sparse shading map, which are reconstructed from the $RGB$ gradient fields over the homogeneous regions, and $N$ denotes a neighbourhood. $E_D$ measures the distance between the final shading map $S(x)$ and the initial (sparse) shading map $S_0(x)$ at pixel $x$, i.e. per-pixel reconstruction accuracy. Then, ${E_S}$ encourages adjacent pixels to have the same shading values, i.e. smoothness.

\section{Intrinsic Image Decomposition}
\begin{figure}[h]
\begin{center}
\includegraphics[width=1\linewidth]{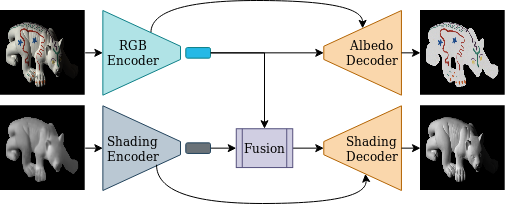}
\end{center}
   \caption{Proposed model architecture. $RGB$ image guides the shading estimation only during the fusion phase using a $1x1$ convolution and a contextual attention module \citep{Yu2018}. Shading decoder only receives shading encoder features through skip connections not to be affected by high resolution $RGB$ color features. Albedo decoder only receives $RGB$ features through skip connections.}
\label{fig:network}
\end{figure}

Since the sparse shading map is completed by only a smoothness constraint, the reconstructed dense map may suffer from geometry loss if the initial gaps are too large. It may also suffer from scale problems due to the least squares fitting. Therefore, we integrate the completed dense shading map into a deep learning framework to refine it and also to predict the corresponding albedo image to achieve intrinsic image decomposition. The network is expected to further improve the shading maps by supervised training and also by the differentiation of additional albedo cues. It is also expected to generate better albedo maps as the dense shading map is robust to color leakages and intensity ambiguities. As stated earlier, deep learning based shading estimations are not as good as albedo estimations. They suffer from albedo color leakages mostly due to texture ambiguities and intensity variations, as demonstrated in Figure~\ref{fig:color_leakage}. On the other hand, our physics-based generated shading map is more robust to those leakages as it is computed only on homogeneous regions. As a result, we design a CNN model such that the $RGB$ image only refines the initial shading estimation, and it is not directly involved in the reconstruction phase to avoid any further critical color leakage. The model is illustrated in Figure~\ref{fig:network}.

\noindent \textbf{Encoders.} Encoder blocks use strided convolution layers for downsampling (4 times). Each convolution is followed by residual blocks~\cite{He2016}. They are preferred as the deviations from the input are rather small. $RGB$ encoder uses 4 consecutive residual blocks, while the shading encoder uses 1 block with different dilation rates. A residual block is composed of Batch Norm-ReLu-Conv(3x3) sequence, repeated twice.\\
\textbf{Fusion.} The final layers of the encoders are fused with a $1x1$ convolution and a contextual attention module \citep{Yu2018} to create a bottleneck such that the related $RGB$ features can properly guide the shading estimation. As a result, the $RGB$ features are fused with the shading features (1) as a (learnable) weighted combination using a $1x1$ convolution, and (2) by the contextual attention module. The contextual attention module learns where to use feature information from known background patches to generate missing patches for the image inpainting task. We adopt their module to our problem such that the shading features uses the information from the $RGB$ features. It is expected to help as in a homogeneously colored patch, the only source causing pixel values to change is the shading component, i.e. $\Delta I = \Delta S$. Therefore, in those regions, the shading map and the $RGB$ image are highly correlated. Fusion happens at $16x16$ resolution. Preliminary experiments suggested that lower resolutions (i.e. $8x8$) cannot reconstruct a decent shading map (too blurry) and higher resolutions (i.e. $32x32$) cause further critical color leakages in the shading.\\
\textbf{Decoders.} The fusion output is fed to the shading decoder, while the albedo decoder takes $RGB$ encoder's final layer as input. Both decoders share the same structure. Encoder features are passed through Conv(3x3)-Batch Norm-LeakyReLu sequence. Then, the feature maps are (bilinearly) up-sampled and concatenated with their encoder counterpart by skip connections. The process is repeated 4 times to reach the final resolution. Shading decoder only receives shading encoder features through skip connections not to be affected by high resolution color features. Albedo decoder only receives $RGB$ features through skip connections. Therefore, we design a specialized network for the intrinsic image decomposition task for robust shading estimation.\\
\textbf{Loss Functions.} The loss functions used to train the model are as follows:
\begin{equation} \label{eq:albedo_loss}
\begin{aligned}
\mathcal{L}_{Albedo} = \lambda_{A1}\;\mathcal{L}_{pixel} + \lambda_{A2}\;\mathcal{L}_{gradient} + \lambda_{A3}\;\mathcal{L}_{dssim} + \lambda_{A4}\;\mathcal{L}_{perceptual}\;,
\end{aligned}
\end{equation}
\begin{equation} \label{eq:shading_loss}
\begin{aligned}
\mathcal{L}_{Shading} = \lambda_{S1}\;\mathcal{L}_{pixel} + \lambda_{S2}\;\mathcal{L}_{gradient} + \lambda_{S3}\;\mathcal{L}_{dssim}\;,
\end{aligned}
\end{equation}
\begin{equation} \label{eq:total_loss}
\begin{aligned}
\mathcal{L}_{Total} = \lambda_{A}\;\mathcal{L}_{Albedo} + \lambda_{S}\;\mathcal{L}_{Shading} + \lambda_{I}\;\mathcal{L}_{Image}\;,
\end{aligned}
\end{equation}
\noindent where $\mathcal{L}_{pixel}$ is the pixel-wise reconstruction loss, which is a weighted combination of mean-squared-error loss and scale-invariant MSE loss, $\mathcal{L}_{gradient}$ denotes the gradient-wise reconstruction loss, $\mathcal{L}_{dssim}$ assesses the structural dissimilarity, $\mathcal{L}_{perceptual}$ measures the reconstruction distance in several feature spaces of a pre-trained VGG16 \citep{Simonyan2015}, $\mathcal{L}_{Image}$  is the image formation loss to force that the estimated reflectance and shading images should reconstruct the original $RGB$ image (i.e. $I = S \times R$), and the $\lambda$s are the weights. Note that the loss functions are the standard reconstruction modules and do not impose any intrinsic image characteristics. The implementation details and other training details are provided in the supplementary material~\footnote{ \url{https://drive.google.com/file/d/1Bl09bJfDS5KayTyHljmHgAmgV5UcxF4E/view?usp=sharing}}.\\\\
\textbf{Dataset.} To train our models, we use the ShapeNet dataset of \citet{Baslamisli2018CVPR}. The dataset includes around 20,000 (synthetic) images of man-made objects randomly sampled from the original ShapeNet dataset \citep{Chang2015}. Following the setup of \citet{Baslamisli2018CVPR}, we render additional images to reach around 50,000 images for training~\footnote{The dataset will be made publicly available.}.

\section{Experiments and Evaluation}
We conduct experiments on four datasets of real world objects with ground-truth intrinsics, MIT Intrinsics~\citep{Grosse2009}, NIR-RGB Intrinsics~\citep{Cheng2019}, Multi-Illuminant Intrinsic Images~\citep{Beigpour2015} and Spectral Intrinsic Images~\citep{Chen2017}. In addition, we provide experiments on two scene-level datasets, As Realistic As Possible~\citep{Bonneel2017} a synthetic ground-truth dataset, and Intrinsic Images in the Wild~\citep{Bell2014} a real world complex dataset with relative human annotations. Finally, we provide further qualitative evaluations on real world in-the-wild images. Comparisons are provided against several state-of-the-art intrinsic image decomposition algorithms. We pick three optimization based methods: (i) STAR, a structure and texture aware advanced Retinex model~\citep{Xu2020}, (ii) IIW, a framework based on clustering and a dense CRF~\citep{Bell2014}, and (iii) SIRFS, a model imposing seven different priors on reflectance, shape and illumination~\citep{Barron2015}. We include four deep learning based methods: (i) ShapeNet uses specialized decoder links to correlate intrinsics and is trained on 2.5M synthetic objects~\citep{Shi2017}, (ii) IntrinsicNet uses deep VGG16 encoder-decoders and an image formation loss, trained on 20K synthetic objects, (iii) RetiNet provides an end-to-end solution to the Color Retinex approach using gradients, trained on 20K synthetic objects, (iv) CGIntrinsics combines two real world scenes (around 3000) and two synthetic scene level datasets (around 20K) for training with additional smoothness constraints to achieve better intrinsics. We use the publicly available models and the original outputs without any fine-tuning or post-processing stages as comparison. To evaluate our proposed method, following the common practice~\citep{Grosse2009}, when dense ground-truths are available, we use the mean squared error (MSE), where the absolute brightness of each image is adjusted by least squares as the ground-truth is only defined up to a scale factor and the local mean squared error (LMSE) with window size 20. For Intrinsic Images in the Wild (IIW) dataset's human annotations, we use Weighted Human Disagreement Rate (WHDR) metric as provided by the authors~\citep{Bell2014}. All the images are resized to $256 \times 256$ for fair comparison. 

\subsection{Evaluations on Object-level Datasets}
 
\subsubsection{MIT Intrinsic Images Dataset}
The dataset contains 20 real-world objects with ground-truth intrinsic images. Objects are lit by a single directional white light source. We follow the recommendation of the authors and exclude apple, pear, phone and potato objects as they are marked as problematic~\citep{Grosse2009}. The quantitative results are provided in Table~\ref{tab:mit}. The table also includes the effect of the contextual attention (CA) module as an ablation study.

 \begin{table}[h]
 \centering
 \scalebox{0.75}{
   \begin{tabular}{|l||c|c|c||c|c|c|}\hline
      \multirow{2}{*}{} &
       \multicolumn{3}{c||}{MSE$~\downarrow$}  &
       \multicolumn{3}{c|}{LMSE$~\downarrow$} \\
       & Shading & Albedo & Average & Shading & Albedo & Average \\ \hline
      STAR & 0.0114 & 0.0137 & 0.0126 & 0.0672 & 0.0614 & 0.0643   \\ \hline 
      SIRFS & \textbf{0.0066} & 0.0129 & 0.0098 & 0.0309 & 0.0572 & 0.0441   \\ \hline 
      IIW & 0.0101 & 0.0210 & 0.0156 & 0.0425 & 0.0720 & 0.0573   \\ \hline \hline 
      ShapeNet & 0.0075 & 0.0158 & 0.0117 & 0.0366 & 0.0543 & 0.0455   \\ \hline
      IntrinsicNet & 0.0304 & 0.0104 & 0.0204 & 0.2038 & 0.0854 & 0.1446  \\ \hline
      RetiNet & 0.0391 & 0.0097 & 0.0244 & 0.2651 & 0.0636 & 0.1644    \\ \hline 
      CGIntrinsics & 0.0117 & 0.0133 & 0.0125 & 0.0425 & 0.0477 & 0.0451    \\ \hline \hline 
      OURS & 0.0069 & \textbf{0.0060} & \textbf{0.0065} & \textbf{0.0418} & \textbf{0.0438} & \textbf{0.0428} \\ \hline 
      OURS (w/o CA) & 0.0075 & 0.0070 & 0.0073 & 0.0454 & 0.0458 & 0.0456 \\ \hline 
   \end{tabular}}
   \caption{Quantitative evaluations on MIT Intrinsic Images dataset. Our proposed model achieves better performance compared against other models on all metrics demonstrating better reconstruction quality. CA module leads to further improvements in performance.} 
   \label{tab:mit}
  \end{table}
  
The results show that comparing with the deep learning based estimations, our proposed models achieves better performance at generating albedo and shading maps on the dataset. Optimization based SIRFS results are better than all other learning based models. Its shading estimations yield the best results on MSE metric. It is known that SIRFS achieves superior performance on single and masked objects, yet it generalize poorly to real scenes~\citep{Narihia2015,Li2018ECCV}. Nonetheless, our estimations are superior than SIRFS on all other metrics. On average, we achieve the best results by a substantial margin. Furthermore, the contextual attention module by~\citet{Yu2018} leads to further performance boost on all metrics. It emerges as a fundamental building block of our proposed method.

In addition, we are extremely efficient compared with the optimization-based methods. To process a single image on MIT dataset, on average, SIRFS takes 111.38 seconds, whereas our model takes 1.79 seconds including the albedo gradient estimation, initial shading recovery from the gradients, filling the initial shading with the smoothness prior, and finally estimating complete intrinsic images. All in all, our model appears 78 times faster than SIRFS. As a side note, IIW model takes 18.09 seconds, and STAR takes 2.78 seconds to process a single image on MIT dataset~\footnote{The results are provided on Intel Xeon CPU E5-2640 v3 @ 2.60GHz.}.

\begin{figure}[h]
\begin{center}
\includegraphics[width=1\linewidth]{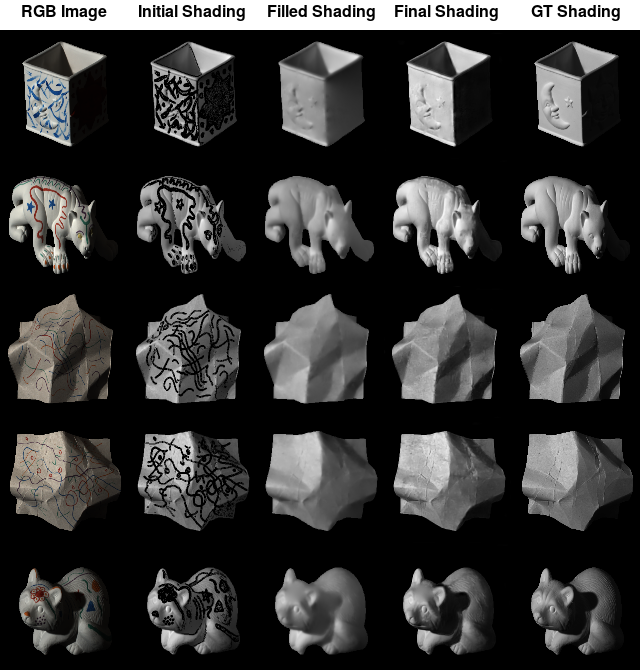}
\end{center}
   \caption{The effect of the proposed framework. The initial shading maps are free of the color leakage problem. The filled shadings are rather blurry, suffers from scale problems and missing geometric details. The deep model further refines it generating sharper shading maps with proper scale.}
\label{fig:prior_effectiveness}
\end{figure}

\begin{figure*}[]
\begin{center}
\includegraphics[width=1\textwidth]{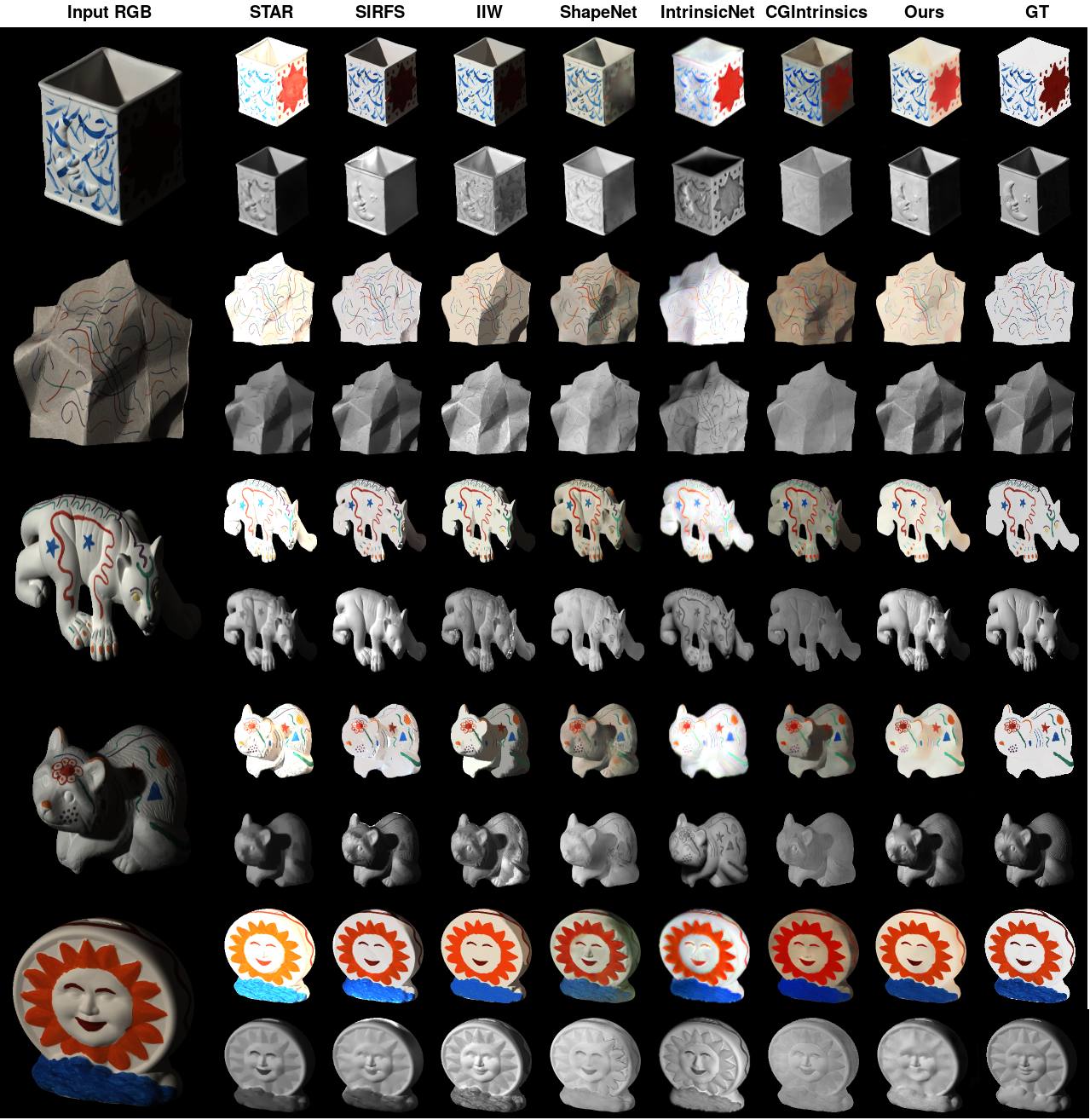}
\end{center}
   \caption{Comparisons with state-of-the-art models. Our shading predictions are more robust to the color leakage problem, while all other methods tend to overfit to the $RGB$ image having severe color leakages in the shading maps. We also achieve significantly better albedo estimations. (Best viewed in digital format.) }
\label{fig:mit_evals}
\end{figure*}

Finally, we provide qualitative evaluations. Figure~\ref{fig:prior_effectiveness} demonstrates the effect of the proposed model from the initial step to reach the final shading map with progressive improvement. The results show that our framework first generates an initial shading map where the color transitions are masked out by the physics-based albedo gradient descriptors. Then the initial shading maps are filled (inpainted/interpolated) with the shading smoothness prior. They are free of color leakages and intensity ambiguities. However, they suffer from scale problems due to the least squares fitting and they are rather blurry due to the neighbourhood smoothness filling. Finally, our deep learning model is able to refine the initially filled shading maps. It makes them sharper, adjusts the scale, and finer geometry details are visible. Figure~\ref{fig:mit_evals} provides the qualitative comparison results against several state-of-the-art models. It shows that we achieve better shadow and shading handling in albedo predictions and our albedo estimations are significantly better. We attribute this to our physics-based shading reconstructions as it handles color leakage and intensity ambiguity problems. Thereby, our shading predictions has no or minimum color leakage. Moreover, the shading map estimations by the deep learning methods tend to severely overfit to the $RGB$ image producing strong color leakages as texture artefacts and intensity ambiguities. 

    \begin{figure}[]
\begin{center}
\includegraphics[width=1\linewidth]{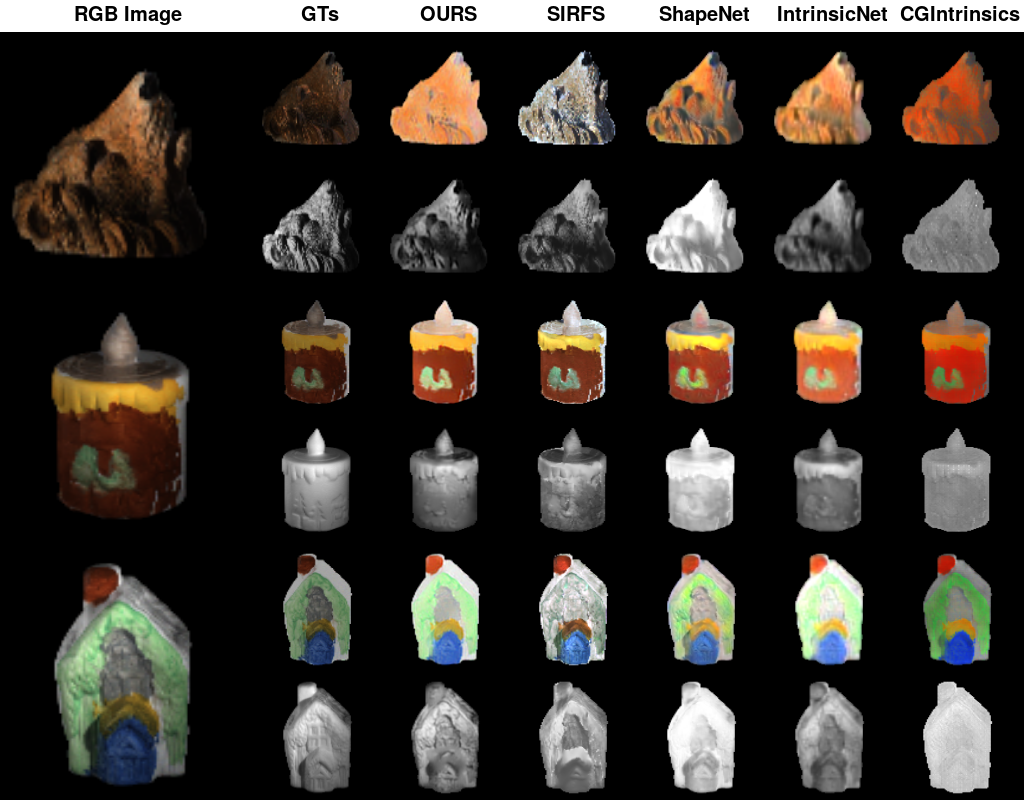}
\end{center}
   \caption{Qualitative evaluations on NIR-RGB Intrinsic Images dataset. Our albedo maps appear more natural and vivid, and closer to the chromaticity patterns of the input images. Our shading estimations do not include intensity ambiguities or texture artefacts.}
\label{fig:nir}
\end{figure}

\subsubsection{NIR-RGB Intrinsic Images Dataset}
We provide additional cross dataset experiments on NIR-RGB Intrinsic Images dataset, which was mainly generated for near-infrared imagery research~\citep{Cheng2019}. It includes seven real-world objects with corresponding ground-truth intrinsics. The quantitative results are provided in Table~\ref{tab:nir}. 
 \begin{table}[h]
 \centering
  \scalebox{0.75}{
   \begin{tabular}{|l||c|c|c||c|c|c|}\hline
      \multirow{2}{*}{} &
       \multicolumn{3}{c||}{MSE$~\downarrow$}  &
       \multicolumn{3}{c|}{LMSE$~\downarrow$} \\
       & Shading & Albedo & Average & Shading & Albedo & Average \\ \hline
      STAR & 0.0028 & 0.0017 & 0.0023 & 0.0896 & 0.1131 & 0.1014   \\ \hline 
      SIRFS & 0.0020 & 0.0009 & 0.0015 & 0.0806 & 0.0950 & 0.0878   \\ \hline 
      IIW & 0.0042 & 0.0018 & 0.0030 & 0.1200 & 0.1345 & 0.1273   \\ \hline \hline 
      ShapeNet & 0.0019 & 0.0008 & 0.0014 & 0.0701 & 0.0772 & 0.0737   \\ \hline
      IntrinsicNet & 0.0021 & 0.0011 & 0.0016 & 0.0748 & 0.0927 & 0.0838  \\ \hline
      RetiNet & 0.0028 & 0.0013 & 0.0021 & 0.0959 & 0.1136 & 0.1048    \\ \hline 
      CGIntrinsics & 0.0027 & 0.0009 & 0.0018 & 0.0862 & 0.0797 & 0.0830    \\ \hline \hline 
      OURS & \textbf{0.0017} & \textbf{0.0006} & \textbf{0.0012} & \textbf{0.0689} & \textbf{0.0609} & \textbf{0.0649}   \\ \hline 
   \end{tabular}}
   \caption{Quantitative evaluations on NIR-RGB Intrinsic Images dataset. Our proposed model achieves better performance compared against other models on all metrics demonstrating better generalization ability.}
   \label{tab:nir}
  \end{table}
  
The results show that our proposed model achieves better performance compared against other models on all metrics. We especially achieve significantly better albedo estimations. The results further demonstrate the improved generalization ability of our proposed method. In this dataset, deep learning based methods are as good as SIRFS, even more superior in some cases. Finally, Figure~\ref{fig:nir} shows qualitative comparisons for a number of images.

The qualitative results further support the quantitative evaluations. Our model predictions are closer to the ground-truth images. The colors of our albedo estimations appear more natural and vivid, and closer to the chromaticity patterns of the input images. Our shading estimations do not include intensity ambiguities or texture artefacts. On the other hand, the intensity ambiguity problem in the shading maps can be observed on ShapeNet and IntrinsicNet estimations on the \emph{candle} and \emph{house} images. CGIntrinsics's shading smoothness constraint tends to generate over-smoothed estimations and cannot capture fine-grained geometric patterns. For example, the balcony of the \emph{house} object is not visible anymore. SIRFS tends to generate incorrect colors on albedo estimations when a scene is dominated by a single color as in the cases of \emph{lion} and \emph{house} objects. The colors of the CGIntrinsics albedo maps tend to shift towards red.

\subsubsection{Multi-Illuminant Intrinsic Images (MIII) Dataset}
MIT Intrinsic Images and NIR-RGB Intrinsic Images datasets provide images with uniform white illumination. 
In this experiment, we further test the ability of our proposed method to generalize also to complex multi-illuminant scenarios. The dataset includes five real-world scenes with multi-colored non-uniform lighting, complex geometry, large specularities, and challenging colored shadows~\citep{Beigpour2015}. Each scene includes two objects and illuminated with 6 single-illuminant and 9 two-illuminants. The colors of the illuminants vary from orange to blue. In total, there are 75 images with ground-truth intrinsics. The quantitative results are provided in Table~\ref{tab:miii}.

The qualitative results show that our proposed model achieves better performance on almost all metrics. Only the reflectance estimations of CGIntrinsics~\citep{Li2018ECCV} are better on the LMSE metric, but their shading estimations are significantly worse. Thus, compared with other works, on average we achieve the best results by a large margin. Note that optimization based SIRFS~\citep{Barron2015} and learning based ShapeNet~\citep{Shi2017} are inherently modelled to estimate multi-colored illumination. Nevertheless, our model emerges more robust to real-world images with multi-colored non-uniform lighting. The results further demonstrate the improved generalization ability of our proposed method. 

 \begin{table}[]
 \centering
   \scalebox{0.75}{
   \begin{tabular}{|l||c|c|c||c|c|c|}\hline
      \multirow{2}{*}{} &
       \multicolumn{3}{c||}{MSE$~\downarrow$}  &
       \multicolumn{3}{c|}{LMSE$~\downarrow$} \\
       & Shading & Albedo & Average & Shading & Albedo & Average \\ \hline
      STAR & 0.0021 & 0.0023 & 0.0022 & 0.0817 & 0.1350 & 0.1084   \\ \hline 
      SIRFS & 0.0003 & 0.0003 & 0.0003 & 0.1015 & 0.1417 & 0.1216   \\ \hline 
      IIW & 0.0003 & 0.0002 & 0.0003 & 0.0869 & 0.1286 & 0.1078   \\ \hline \hline 
      ShapeNet & 0.0002 & 0.0002 & 0.0002 & 0.0846 & 0.1020 & 0.0933   \\ \hline
      IntrinsicNet & 0.0002 & 0.0002 & 0.0002 & 0.0597 & 0.0873 & 0.0735  \\ \hline
      RetiNet & 0.0002 & 0.0002 & 0.0002 & 0.0590 & 0.0964 & 0.0777    \\ \hline 
      CGIntrinsics & 0.0004 & 0.0001 & 0.0003 & 0.1172 & \textbf{0.0707} & 0.0940    \\ \hline \hline 
      OURS & \textbf{0.0002} & \textbf{0.0001} & \textbf{0.0002} & \textbf{0.0514} & 0.0770 & \textbf{0.0642}   \\ \hline 
   \end{tabular}}
   \caption{Quantitative evaluations on MIII dataset with multi-colored non-uniform lighting. Our proposed model achieves better performance and is more robust to multi-colored non-uniform lighting.}
   \label{tab:miii}
  \end{table}
  
\subsubsection{Spectral Intrinsic Images Dataset (SIID)}
The dataset was mainly generated for spectral intrinsic image decomposition research~\citep{Chen2017}. It includes nine objects illuminated with two kinds of light sources, one white and one warm-tone white. In total, it has 18 spectral images with corresponding shading ground-truths. The dataset also provides corresponding $RGB$ images synthesized from the spectral images that are used as inputs to the models. The quantitative results are provided in Table~\ref{tab:siid}. 

\begin{table}[h]
   \centering
   \begin{tabular}{|c||c|c|}\hline
      & MSE-s$~\downarrow$ & LMSE-s$~\downarrow$ \\ \hline
     STAR & 0.0034 & 0.0192 \\ \hline 
     SIRFS & 0.0186 & 0.0215 \\ \hline 
     IIW & 0.0064 & 0.0164   \\ \hline \hline 
     ShapeNet & 0.0129 & 0.0424   \\ \hline 
     IntrinsicNet & 0.0045 & 0.0189   \\ \hline 
     RetiNet & 0.0047 & 0.0220   \\ \hline 
     CGIntrinsics & 0.0142 & 0.0286   \\ \hline \hline 
     OURS & \textbf{0.0027} & \textbf{0.0156}   \\ \hline 
   \end{tabular}
   \caption {Quantitative evaluations on SIID dataset with white and warm-tone white illuminations. Our proposed model achieves better performance and has better generalization ability.}
   \label{tab:siid}
 \end{table}

\begin{figure}[t]
\begin{center}
\includegraphics[width=1\linewidth]{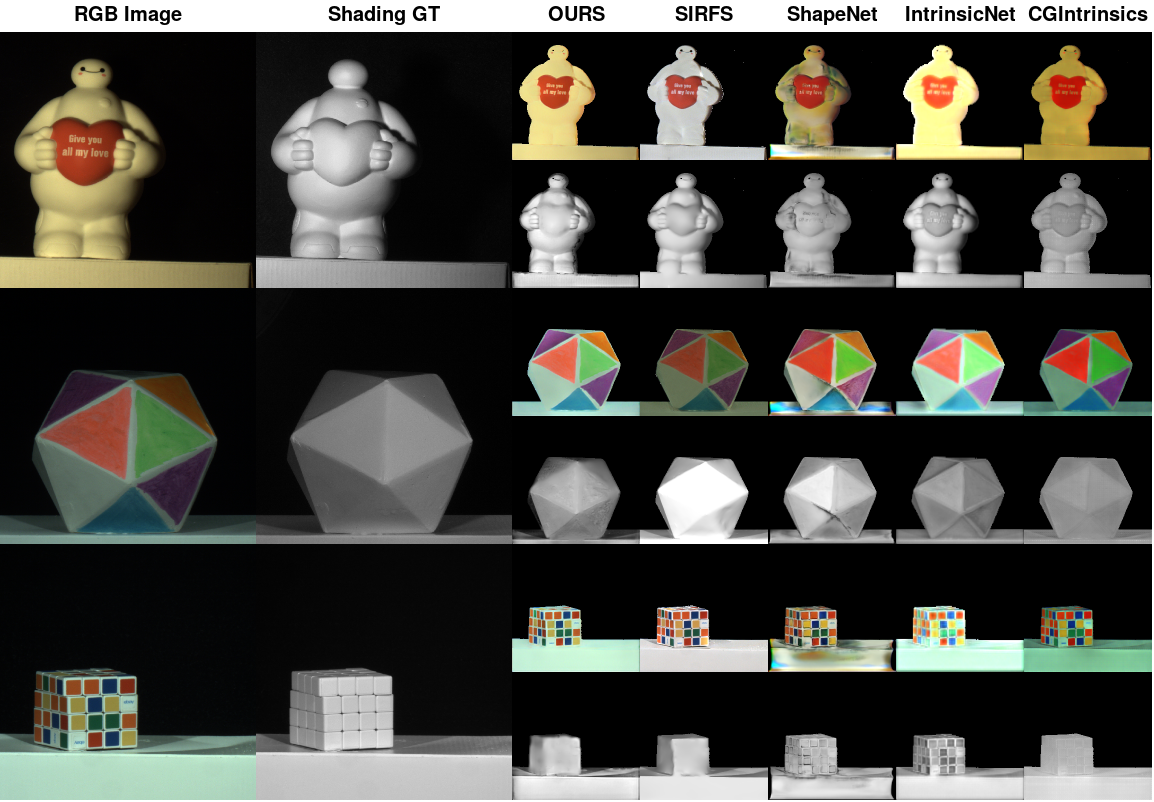}
\end{center}
   \caption{Qualitative evaluations on SIID. Our albedo maps appear more natural and vivid. Our shading estimations do not include intensity ambiguities or texture artefacts and are closer to the ground-truths.}
\label{fig:siid}
\end{figure}

The results show that the reconstruction quality of our shading maps are closer to the ground-truths on all metrics. Similar to the MIII dataset experiments with multi-colored non-uniform lighting, our models also achieve more robust results on a different illumination setting of warm-tone white. Finally, Figure~\ref{fig:siid} shows qualitative comparisons for a number of images.

The qualitative results further support the quantitative evaluations. Our model predictions are closer to the ground-truth images. Our albedo estimations appear more natural and vivid and they are free of geometric effects. Our model is also capable of removing shadows casts on the platforms of the \emph{gypsum} and \emph{cube} objects from the albedo estimations. Since our model is trained only on white light, the color of the light source is also embedded in the albedo. Same behaviour is also observed on other models. To overcome this issue, a white balancing algorithm can be applied to the input images as a pre-processing step. Nonetheless, it does not cause significant problems on the reconstruction quality as the ground-truths are not absolute and only defined up to a scale factor~\citep{Grosse2009,Narihia2015}. SIRFS can handle the issue, but it tends to confuse albedo and color of the light source when a scene is dominated by a single color as demonstrated in the previous section. Additional examples can be found in the upcoming sections. Likewise, as mentioned in the previous section, ShapeNet~\citep{Shi2017} is inherently modelled to estimate multi-colored illumination. However, it also fails to differentiate the color of the light source and albedo in this case. It also generates undesired color artefacts on the albedo maps.

As for the shading map generations, our model estimations are free of any texture artefacts and intensity ambiguities. The text on the heart of the \emph{baymax} object is correctly attributed to the albedo map, whereas ShapeNet estimation is contaminated with the texture artefact, and IntrinsicNet and CGIntrinsics estimations both contain texture artefacts and intensity ambiguities. The intensity ambiguity problem is more severe on the shading estimations of the \emph{cube} object. Our model and optimization-based SIRFS can handle those. Nevertheless, our contribution is more significant on the \emph{gypsum} object, where SIRFS tend to generate over-smooth and overly-bright estimations that the geometry is distorted and fine-grained structures are not visible anymore. Our model is also not flawless. For example, we cannot capture the fine geometric details of the \emph{cube} image and our estimation appears more rigid. That is because of the shading smoothness constraint that is used to fill in the gaps of the initial shading map based on the neighboring pixel information. Since the color changes happen near the holes, shading smoothness interpolation also fills in those gaps. Therefore, the shading estimation appears more rigid in those cases.

\subsubsection{Amsterdam Library of Object Images (ALOI) Dataset}
\begin{figure}[t]
\begin{center}
\includegraphics[width=1\linewidth]{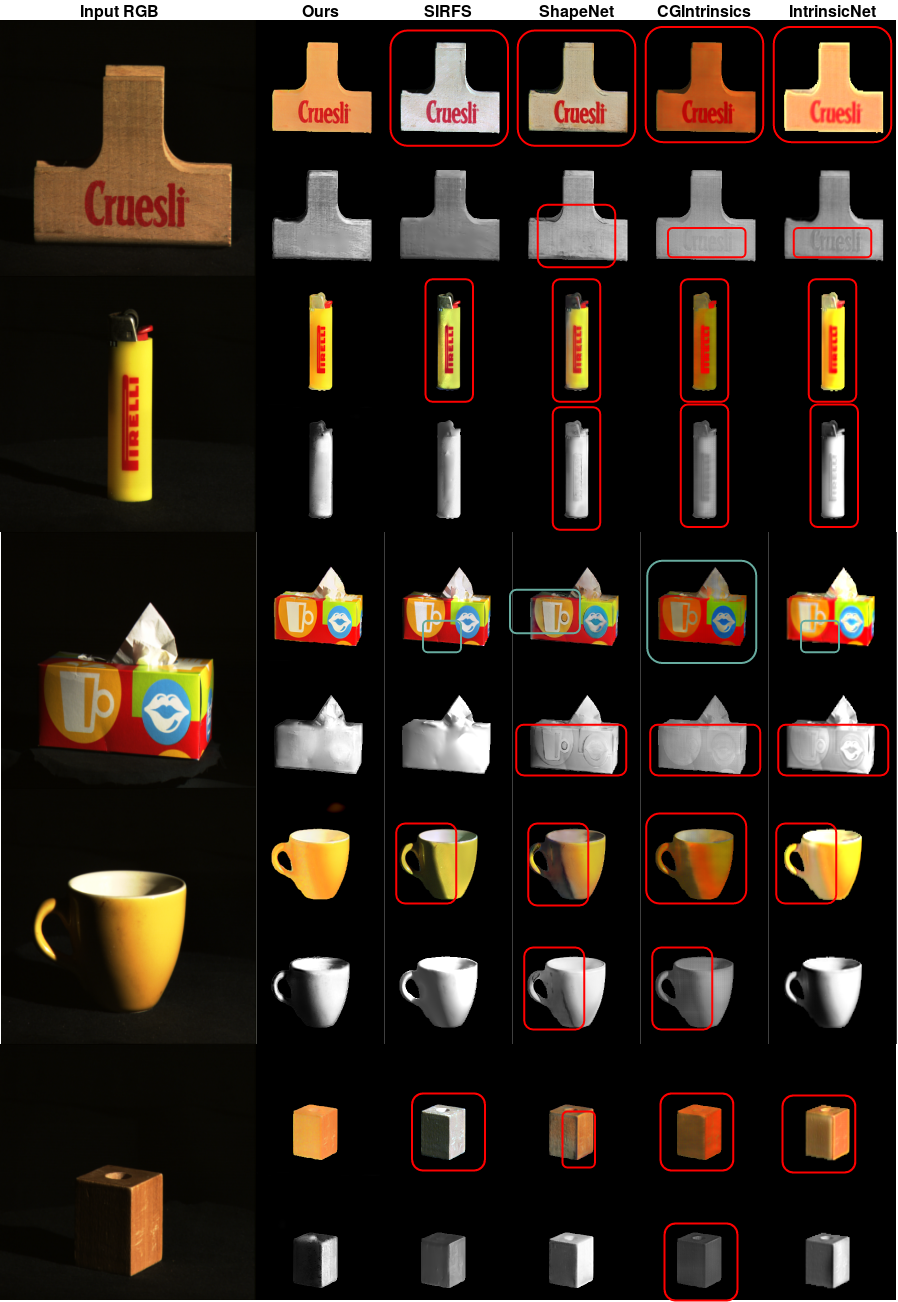}
\end{center}
   \caption{Additional real world evaluations on ALOI dataset. Rows (1,2,3) provide examples with textures,  and (4,5) with strong shading patterns. Deep learning methods have severe color leakages in the shading maps and cannot handle strong shadings in the albedo maps. Our method is capable of capturing decent albedo and shading maps for also ALOI images.}
\label{fig:aloi}
\end{figure}

We also provide additional visual comparisons for real world images without ground-truths. For the task, we use Amsterdam Library of Object Images (ALOI) dataset~\citep{Geusebroek2005}. Figure~\ref{fig:aloi} provides a number of example objects with different properties to demonstrate the effectiveness of our method. Rows (1,2,3) provide examples with textures and rows (4,5) provides examples with strong shading patterns.

Deep learning based methods have severe color leakages in the shading map estimations for textured objects. CGIntrinsics's shading smoothness constrain negatively effects the shading maps when strong shading patterns are present. It generates homogeneously smooth images such that it cannot properly capture darker regions where the surface normals (geometry) significantly deviate from the incoming light source direction. It can be observed from the \emph{cup} image that the right part of the handle should be covered by the shading pattern and should not be visible. Our proposed work is the only model that can capture that pattern. Similar behaviour is also observed for the \emph{wooden cube} in the last row. Likewise, the other models cannot generate a decent albedo map in those cases. ShapeNet generated albedo maps are rather dull colored and blurry. Similarly, CGIntrinsics and IntrinsicNet generated albedo maps tend to be polluted with color artefacts. On the other hand, our model is better at avoiding attributing surface texture to the shading maps, and our albedo estimations are sharper, have better color augmentation and more natural for all cases. SIRFS model is capable of producing decent shading maps for textured objects, as well. However, its albedo predictions are not as decent when an image is dominated by a single color as in the case of 1st and 5th rows. Similarly, it tends to fail to capture decent shading maps when an image has strong shading patterns. 

\subsection{Evaluations on Scene-level Datasets}
There are several aspects that are challenging for our current setup for the scene level intrinsic image decomposition. Firstly, a scene is composed of multiple objects so that the behaviour of the illumination component is more complex. Especially, the ambient light (inter-reflection) effect is way stronger. In addition, our optimization process using the smoothness constraint to fill in the gaps of the initial shading map may be negatively effected if the gaps are filled from different surfaces (e.g. filled with object boundaries). Similarly, cluttered objects may cause way too large gaps to fill. Another thing is that since scene level objects have different scales, one single threshold might not be sufficient to obtain proper gradients. Nonetheless, for the sake of completeness, we also evaluate our model on scene-level images to provide additional insights.
 
\subsubsection{As Realistic As Possible (ARAP) Dataset}
With the current technology, it is not possible to generate dense ground-truth intrinsic images for any real world scene. Collecting the ground-truth intrinsics happens only on object-level and in a fully-controlled (indoor) laboratory settings, which demands extreme care~\citep{Grosse2009,Chen2017,Cheng2019}. That is the reason why those datasets are small sampled. Therefore, to evaluate our model on scene-level images, we utilize the synthetic dataset of~\citet{Bonneel2017}. The dataset provides 53 high quality realistic scene-level renderings with corresponding per-pixel ground-truth intrinsics. Some of the scenes were re-rendered with different illumination settings. Thus, the evaluation is provided for the full dataset of 152 images. The quantitative results are provided in Table~\ref{tab:arap}. The evaluations do not include CGIntrinsics model as it uses ARAP for training~\citep{Li2018ECCV}, and also SIRFS model as it is specifically designed for single objects and generalize poorly to real scenes~\citep{Narihia2015,Li2018ECCV}. Compared with other frameworks our proposed model achieves better performance on all metrics also on scene-level images, which further demonstrates our improved generalization ability. 

 \begin{table}[h]
 \centering
    \scalebox{0.75}{
   \begin{tabular}{|l||c|c|c||c|c|c|}\hline
      \multirow{2}{*}{} &
       \multicolumn{3}{c||}{MSE$~\downarrow$}  &
       \multicolumn{3}{c|}{LMSE$~\downarrow$} \\
       & Shading & Albedo & Average & Shading & Albedo & Average \\ \hline
      IIW & 0.0913 & 0.0496 & 0.0705 & 0.2050 & 0.0721 & 0.1386   \\ \hline \hline 
      ShapeNet & 0.1218 & 0.0978 & 0.1098 & 0.2400 & 0.1435 & 0.1918   \\ \hline
      IntrinsicNet & 0.0889 & 0.0380 & 0.0635 & 0.1867 & 0.0530 & 0.1199  \\ \hline
      RetiNet & 0.0874 & 0.0417 & 0.0646 & 0.1875 & 0.0600 & 0.1238    \\ \hline 
      OURS & \textbf{0.0862} & \textbf{0.0337} & \textbf{0.0600} & \textbf{0.1832} & \textbf{0.0482} & \textbf{0.1157}   \\ \hline 
   \end{tabular}}
   \caption{Quantitative evaluations on scene-level ARAP dataset. Our proposed model achieves better performance and generalization ability.}
   \label{tab:arap}
  \end{table}

\subsubsection{Intrinsic Images in the Wild (IIW) Dataset}
We follow the common practice and utilize the test set used by previous work~\citep{Zhou2015,Li2018ECCV}. The test split includes 1046 images with relative human annotations. The quantitative results are provided in Table~\ref{tab:iiw}. We also train our model with less data (20K) to provide a more fair comparison against the models of ~\citet{Baslamisli2018CVPR}.

\begin{table}[h]
   \centering
   \begin{tabular}{|c||c|c|}\hline
      & Training set & WHDR $~\downarrow$ \\ \hline
     STAR & - & 32.9\%  \\ \hline 
     IIW & - & 20.6\%    \\ \hline \hline 
     DirectIntrinsics & Sintel & 37.3\%   \\ \hline 
     CGIntrinsics & SUNCG & 26.1\%    \\ \hline 
     CGIntrinsics & CGI & 18.4\%    \\ \hline \hline 
     ShapeNet & ShapeNet (2.5 M) & 59.4\%    \\ \hline 
     IntrinsicNet & ShapeNet (50 K) & 32.1\%    \\ \hline 
     RetiNet & ShapeNet (50 K) & 37.9\%    \\ \hline \hline 
     OURS & ShapeNet (20 K) & 28.7\%    \\ \hline 
     OURS & ShapeNet (50 K) & 28.9\%    \\ \hline 
     OURS* & ShapeNet (50 K) & 26.8\%    \\ \hline 
   \end{tabular}
   \caption {Quantitative evaluations on IIW dataset with human annotations. Our proposed model achieve significantly better reflectance predictions among the models trained on object-level ShapeNet dataset. ${*}$ indicates that the CNN predictions are post-processed with a guided filter.}
   \label{tab:iiw}
 \end{table}

\begin{figure}[!h]
\begin{center}
\includegraphics[width=1\linewidth]{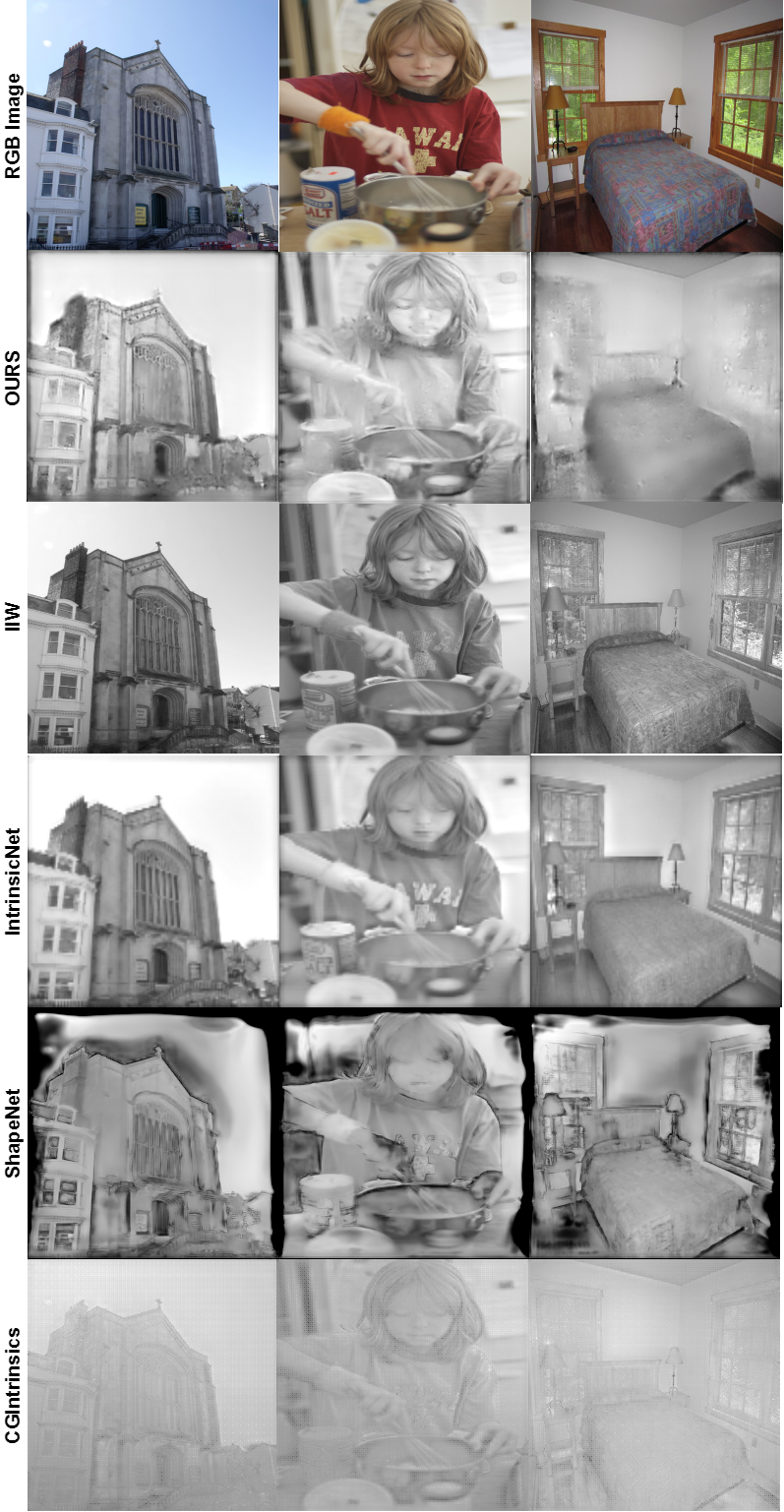}
\end{center}
   \caption{Shading evaluations on IIW. Our model can produce scene-level shading maps that are free of texture or intensity ambiguities. Other models tend to overfit to the RGB images.}
\label{fig:iiw_shading}
\end{figure}

\begin{figure}[!t]
\begin{center}
\includegraphics[width=1\linewidth]{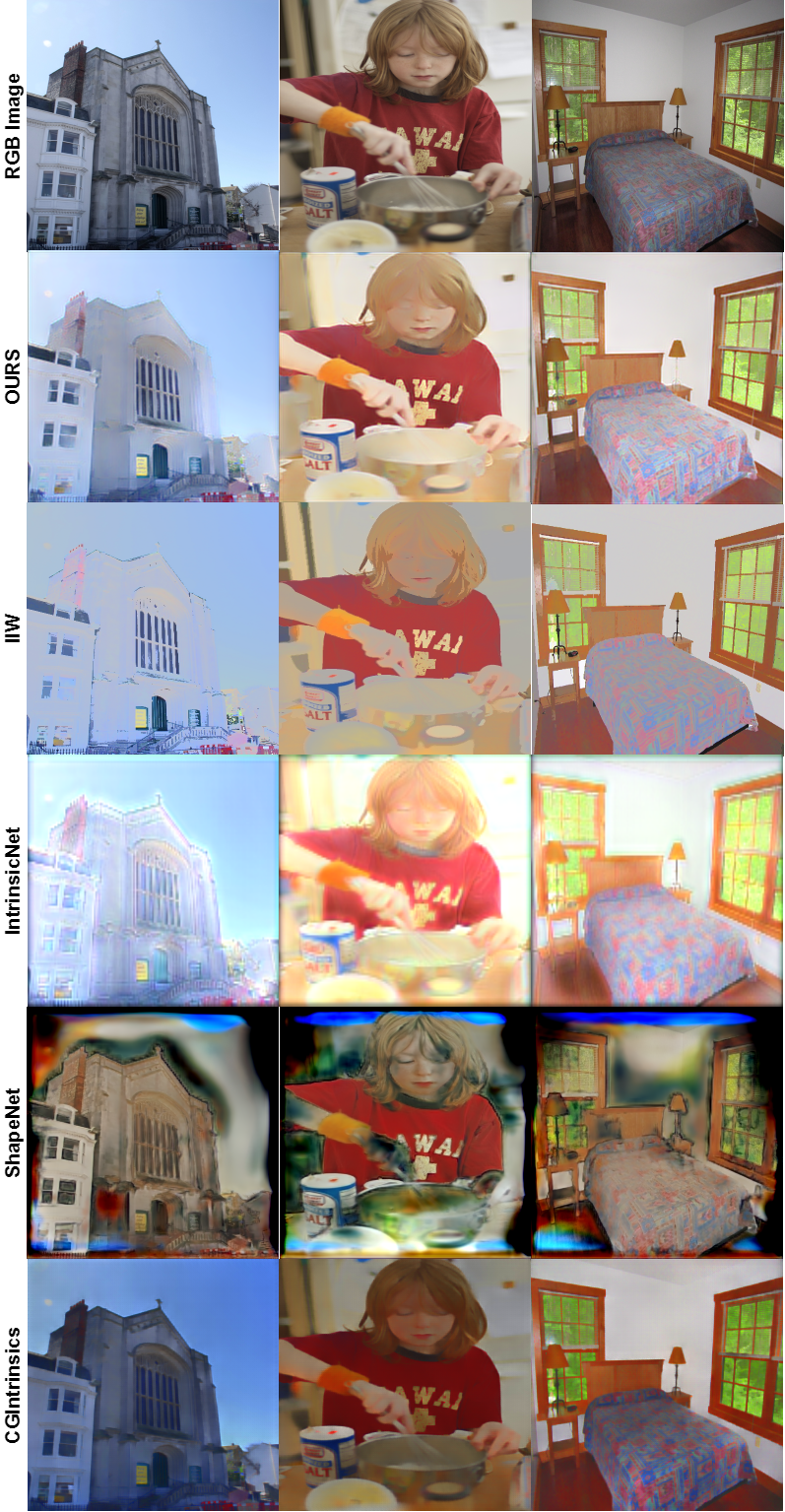}
\end{center}
   \caption{Albedo evaluations on IIW. Our model can generate proper scene-level albedo maps. We can also handle strong shadow casts.}
\label{fig:iiw_albedo}
\end{figure}

Comparing with the models trained on object-level ShapeNet dataset, our proposed model achieve significantly better reflectance predictions. Additional performance boost is achieved by applying a post processing step to enforce piecewise constant reflectance~\citep{Nestmeyer2017}. Decreasing the training sample size does not significantly effect the performance for our model's albedo estimations on IIW. Furthermore, our proposed model is significantly better than the structure and texture aware advanced Retinex model, and also DirectIntrinsics model trained on scene-level Sintel dataset. We also achieve on par results with CGIntrinsics model when trained on scene-level SUNCG dataset. The model achieves superior performance by combining the refined and improved renderings of scene-level SUNCG and the integration of ARAP dataset to create their final dataset CGI. It is also worthwhile to note that all the learning based models use data augmentations through random flips, shifts, resizings, and crops, whereas we do not apply any augmentation technique. Finally, Figure~\ref{fig:iiw_albedo} provides qualitative comparisons for albedo estimations, and Figure~\ref{fig:iiw_shading} for shading estimations.

ShapeNet estimations are contaminated with artefacts and do not appear natural. The shading of the \emph{bed} image includes texture artefacts and the text \emph{AWAI} is directly copied to the shading map in the \emph{girl} image. Similar patterns are also observed in IntrinsicNet estimations. IntrinsicNet generated shading maps also suffer from intensity ambiguities, which can be observed from the \emph{girl} image that the neck of of the t-shirt has a darker color. Its albedo estimations are better than ShapeNet's, yet they contain inconvenient brightness artefacts. IIW's albedo estimations appear natural and free of geometry effects. However, its shading generations directly overfit to the $RGB$ inputs, and all the texture patterns are clearly visible in the shading maps. 

\noindent CGIntrinsics trained on scene-level imagery achieves decent albedo predictions with proper smoothing effects, and compared with others, they appear more natural. However, their shading estimations appear way too smooth and hazy and most of the structures are not visible anymore, see the stairs or the fine-grained pillars of the church. It also suffers from the same intensity ambiguity problem as IntrinsicNet. On the other hand, our model is also capable of producing scene-level shading maps that are free of texture or intensity ambiguities. The first image shows that our model also works on outdoor scenes capable of handling geometry differences and different light properties. We can also handle the text on the t-shirt of the \emph{girl} image and the text on the \emph{salt box} and correctly attribute them to albedo maps. The windows of the \emph{bed} image are an example where our shading map is negatively effected as our model tries fill in the gaps with insufficient gradient information. Although we did not enforce it as CGIntrinsics, our albedo estimations also appear smooth. However, our method still makes mistakes, such as the face of the \emph{girl} or right side of the \emph{church} appear blurry. Finally, our model is the only one that can handle the strong shadow cast under the bed. Our albedo estimations are free of strong shadow casts in this example, whereas all other models fail to handle it.

\section{Conclusion}
We investigated the use of photometric invariance to steer a deep learning model for intrinsic image decomposition (albedo and shading). We proposed albedo and shading gradient descriptors which are derived from physics-based models. Using the descriptors, albedo transitions are masked out and an initial shading map is calculated directly from the corresponding $RGB$ image gradients in a learning-free unsupervised manner. Then, an optimization method was proposed to reconstruct the full dense shading map. Finally, we integrated the generated shading map into a novel deep learning framework to refine it and also to predict corresponding albedo image to achieve intrinsic image decomposition. Additionally, to train our model, a large-scale dataset of synthetic images of man-made objects was extended from 20K to 50K.

The evaluations were provided on five different object-level datasets (MIT, NIR-RGB, MIII, SIID, and ALOI), and two scene-level datasets (ARAP and IIW) with comprehensive setups without any fine-tuning or domain adaptation stage. The evaluations proved that our proposed model generated shading maps are more robust to texture artefacts and intensity ambiguities, which has been a long standing problem in the intrinsic image decomposition task. Since our model handles the undesired artefacts in the shading estimations, we also better differentiate albedo changes and achieve superior quantitative results.

Another conclusion is that deep learning based methods tend to overfit to the $RGB$ image having critical color leakages in the shading maps. When quantitatively evaluating, the leakage effect may not be reflected numerically. That suggests that future work should focus on proposing better metrics for evaluation. In addition, the color leakage effect may not be observed when a model is trained and tested (or fine-tuned) on the same dataset~\citep{Narihia2015,Cheng2018}. Therefore, it is important for intrinsic image decomposition methods to provide cross-dataset or in-the-wild evaluations. Finally, we also tried to adapt several guided image-to-image translation and feature modulation techniques for our preliminary experiments to refine our initial shading maps with the $RGB$ features. In particular, we tried the end-to-end trainable guided filter by~\citet{Wu2018}, bi-directional guided image-to-image translation by~\citet{AlBahar2019}, spatially-adaptive normalization by~\citet{Park2019}, and deep spatial feature transform by~\citet{Wang2018}. Unfortunately, none of them were able to address the color leakage problem in the shading maps. 

\section*{Acknowledgments}
This project was funded by the EU Horizon 2020 program No. 688007 (TrimBot2020). We thank Partha Das for his contribution to the experiments.
%Acknowledgments should be inserted at the end of the paper, before the
%references, not as a footnote to the title. Use the unnumbered
%Acknowledgements Head style for the Acknowledgments heading.

\bibliographystyle{model2-names}
\bibliography{refs}

\end{document}